\documentclass[11pt]{article}

\usepackage[margin=1in]{geometry}

\usepackage[T1]{fontenc}
\usepackage[utf8]{inputenc}
\usepackage{lmodern}
\usepackage{microtype}

\usepackage{amsmath,amssymb}

\usepackage{graphicx}
\usepackage{caption}
\usepackage[round,numbers,sort&compress]{natbib}

\usepackage{authblk}

\usepackage{lineno}
\usepackage{xcolor}
\usepackage{hyperref}
\hypersetup{
    colorlinks=true,
    linkcolor=blue!60!black,
    citecolor=blue!60!black,
    urlcolor=blue!60!black,
}

\title{\textbf{Machine individuality: Separating genuine idiosyncrasy from response bias in large language models}}

\author[1,*]{Valentin Kriegmair}
\author[1,2]{Dirk U. Wulff}
\affil[1]{Center for Adaptive Rationality, Max Planck Institute for Human Development, 14195 Berlin, Germany}
\affil[2]{Center for Cognitive and Decision Sciences, Department of Psychology, University of Basel, 4055 Basel, Switzerland}
\affil[*]{Corresponding author (kriegmair@mpib-berlin.mpg.de)}
\date{\today}

\begin{document}
\maketitle

% --- Abstract ---
\begin{abstract}
As large language models (LLMs) are increasingly integrated into daily life, in roles ranging from high-stakes decision support to companionship, understanding their behavioral dispositions becomes critical. A growing literature uses psychometric inventories and cognitive paradigms to profile LLM dispositions. However, these approaches cannot determine whether behavioral differences reflect stable, stimulus-specific individuality or global response biases and stochastic noise. Here, we apply crossed random-effects models---widely used in psychometrics to separate systematic effects---to 74.9 million ratings provided by 10 open-weight LLMs for over 100,000 words across 14 psycholinguistic norms. On average, 16.9\% of variance is attributable to stimulus-specific individuality, robustly exceeding a statistical null model. Cross-norm prediction analyses reveal this individuality as a coherent fingerprint, unique to each model. These results identify individual differences among LLMs that cannot be attributed to response biases or stochastic noise. We term these differences \emph{machine individuality}.
\vspace{0.5em}

\end{abstract}

\vspace{1em}

As large language models (LLMs) are deployed for a widening range of purposes, from high-stakes decision support to everyday companionship, understanding their behavioral dispositions becomes consequential. Whether a model renders moral judgments harshly or gently, or rates emotional content vividly or flatly, shapes its usability and performance. Acknowledging this point, major providers now offer models with distinct personality modes. A growing body of work has begun to address the issue by applying psychometric inventories~\citep{serapio2023personality, pellert2024aipsychometrics}, cognitive paradigms~\citep{binz2023turning, hagendorff2023machine}, and classification methods~\citep{sun2025idiosyncrasieslargelanguagemodels} to characterize the behavioral dispositions of LLMs. These efforts share an implicit assumption---that LLM behavior varies in stable, measurable ways across models---for which evidence remains surprisingly weak. Measured ``personality'' scores shift substantially with prompt template~\citep{sclar2024quantifying}, option ordering~\citep{gupta2024selfassessment}, and scale type~\citep{li2025decoding}. Here, we ask a more fundamental question: Does stable behavioral individuality---separable from shared consensus, response biases, and stochastic noise---exist in LLMs at all?

This requires separating sources of variation that existing approaches conflate---a challenge personality psychology has long recognized. Behavioral differences can reflect substantive traits (stable, stimulus-specific dispositions) or response styles such as acquiescence and extreme responding that shift all responses directionally~\citep{paulhus1991measurement, wetzel2016response}. Yet individuality resides in person-by-situation interactions. The same ambiguity afflicts current characterizations of LLM behavior. Most compellingly, Sun et al.~\citep{sun2025idiosyncrasieslargelanguagemodels} show that a fine-tuned classifier identifies source LLMs with 97\% accuracy, even after paraphrasing. Yet their own analyses reveal that the discriminative signal is carried predominantly by surface-level stylistic regularities, such as characteristic phrases or formatting conventions, features that function as the machine analogue of response styles. In short, existing approaches can tell models apart but cannot determine whether the differences reflect situation-specific evaluation or general response styles.

\begin{figure}[!htbp]
\centering
\includegraphics[width=\linewidth]{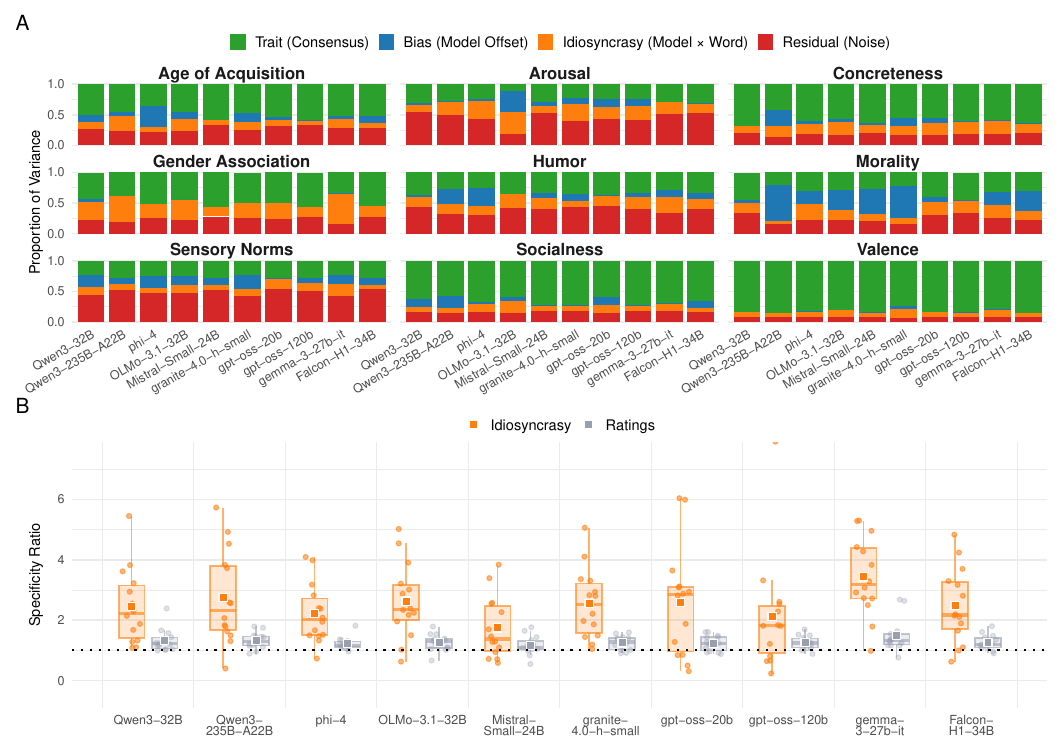}
\caption{Machine Individuality: Variance Partitioning and Specificity.
\textbf{A.} Crossed random-effects decomposition of $\sim$5.4M ratings per norm (10 open-weight LLMs, 5 repetitions) across 14 psycholinguistic norms (five sensory norms and two Age of Acquisition norms grouped for brevity). Variance decomposes into Trait (consensus), Bias (model offset), Idiosyncrasy (Model $\times$ Word), and Residual. Idiosyncrasy averages 16.9\% (4.8\%--34.0\% across norms).
\textbf{B.} Ridge regression specificity ratios: Each model's per-word idiosyncrasy estimates on a held-out norm predicted from the remaining 13, relative to cross-model baselines. All ratios exceed 1.0 on average (1.74--3.43), confirming coherent, model-specific semantic fingerprints.}
\label{fig:main}
\end{figure}

To untangle situation-specific individuality from directional bias, we turn to crossed random-effects models~\citep{bates2015lme4}, the standard psychometric tool for decomposing observed variation into systematic effects, such as shared consensus, systematic biases, and the person-by-situation interaction that constitutes individuality. This decomposition requires a fully crossed design---every respondent responds to every situation. Standard psychometric inventories are ill-suited: They were designed for human self-report, and their construct validity does not transfer to LLMs~\citep{suhr2025stop}. What would constitute a valid behavioral assessment for LLM individuality is far from obvious. Here, we propose psycholinguistic word ratings---numerical judgments along dimensions such as valence, arousal, concreteness, or morality---as a novel approach. These ratings were obtained for over 100,000 words across virtually the entire semantic space. By rating this broad lexicon, a model effectively reveals how it would evaluate virtually any situation. We administered zero-shot rating tasks to 10 open-weight LLMs spanning a broad range of architectural implementations and parameter scales (20B--235B), eliciting ratings across 14 psycholinguistic norms (see Figure~\ref{fig:main}A) via repeated stochastic sampling ($\sim$74.9 million total observations). A crossed random-effects model then decomposed each rating $y_{ijk}$ (word $i$, model $j$, repetition $k$):
\begin{equation}
 y_{ijk} = \mu + \tau_i + \beta_j + \iota_{ij} + \epsilon_{ijk}
\end{equation}
into shared \emph{Trait} ($\tau_i$), systematic model \emph{Bias} ($\beta_j$), \emph{Residual} noise ($\epsilon_{ijk}$), and---critically---model \emph{Idiosyncrasy} ($\iota_{ij}$). This interaction captures how each model evaluates each word, net of consensus and bias. If it accounts for zero variance, apparent behavioral differences reduce to Bias and Residual; conversely, if it is substantial, and larger than Bias, models possess genuine individuality.

\section*{Results}

Figure~\ref{fig:main}A displays the crossed random-effects decomposition of $\sim$5.4M ratings per norm across 14 psycholinguistic norms, with the five sensory norms (Auditory, Gustatory, Haptic, Olfactory, and Visual) grouped. Shared Trait variance dominates (32.1\%), confirming broad cross-model consensus on the semantic landscape~\citep{jiang2025artificial}. The critical question is how the remaining variance partitions between systematic Bias and stimulus-specific Idiosyncrasy. The answer is unambiguous: Idiosyncrasy accounts for 16.9\% of total variance on average, ranging from 4.8\% (Valence) to 14.3\% (across sensory norms; 9.7\%--18.0\% individually) to 34.0\% (Gender Association), and robustly exceeds parametric null expectations across all 14 norms (null upper bound $< 0.033\%$; all $p < 0.01$). Models differ not only in their general response tendencies, but in how they evaluate specific words.

Crucially, these idiosyncrasies are not isolated to individual dimensions but form internally coherent, model-specific fingerprints. Using Ridge regression to predict each model's best linear unbiased predictions (BLUPs)---i.e., the estimated per-word deviations $\hat{\iota}_{ij}$ after removing shared trait and bias effects---on a held-out norm from its BLUPs on the remaining 13 norms, we found that the mean specificity ratio for every model is larger than $1.0$ (Fig.~\ref{fig:main}B; range: 1.74--3.43), and exceeds that of the model ratings (range: 1.16--1.49). This indicates that a model's deviations on one norm are best predicted by its own deviations on other norms rather than by those of other models.

The structure of idiosyncrasy varies across norms. Gender Association and Arousal show the strongest variation (34.0\% and 22.6\%), while Gustatory ratings show the least (9.7\%); Morality is distinguished by substantial Bias; and Valence, Socialness, Concreteness, and Age of Acquisition are dominated by shared Trait variance, reflecting strong cross-model consensus. Models correlated only moderately with human norms ($\bar{r} = 0.48$--$0.67$), suggesting LLM semantic representations are partly sui generis rather than noisy approximations of human judgments. Stochastic aggregation consistently outperformed deterministic decoding in predicting human judgments ($\Delta\bar{r} = 0.032$; 10/10 models, 14/14 norms), introducing a reproducibility--alignment tradeoff: Deterministic decoding maximizes replicability but sacrifices both variance structure and human alignment.

\section*{Conclusion}

Across 14 psycholinguistic norms, model idiosyncrasy accounts for a substantial share of behavioral variance (4.8--34.0\%) beyond shared semantic consensus and systematic directional bias. These stimulus-specific deviations form coherent, cross-dimensional fingerprints---what we term \emph{machine individuality}.

This result has direct implications for how LLM dispositions are evaluated. Just as personality psychologists learned to separate substantive traits from response styles~\citep{paulhus1991measurement}, behavioral evaluations of LLMs must separate stimulus-specific individuality from directional bias. Our decomposition shows that the two coexist in substantial proportions, varying sharply across psychological dimensions---a distinction invisible to aggregate personality scores. Yet without it, claims about model character are fundamentally ambiguous: It remains unknown whether they reflect how a model evaluates situations or merely how it tends to respond.

While psycholinguistic word ratings proved a powerful paradigm for isolating individuality---spanning the breadth of semantic space across evaluative dimensions including morality, valence, and arousal---the decomposition framework itself has general applicability: It applies wherever multiple models respond to a shared set of stimuli. Open questions are whether these idiosyncrasies persist under contextualized elicitation~\citep{kumar2026failure} and whether they predict downstream behavioral differences. Evidence for the latter would establish that machine individuality is not only measurable but consequential.

\section*{Methods}

We selected 10 open-weight LLMs (20B--235B parameters; full list in \emph{SI Appendix}) and curated 107,083 cues from the \texttt{psychNorms} metabase~\citep{hussain2024probingcontentssemanticrepresentations}, spanning 14 psycholinguistic norms (see \emph{SI}). Using zero-shot prompts adapted from the original studies' scale descriptions, we elicited ratings via stochastic sampling ($T = 1.0$, 5 repetitions per item) and deterministic decoding ($T = 0$). Each prompt placed the model in a rating task, presented the original study's scale anchors, and constrained responses to a single numeric value (see \emph{SI Appendix} for full templates). After exclusion of invalid, outlier responses ($<3\%$ missing per norm, mean $<0.5\%$), approximately 5.4~million observations per norm entered the analysis (${\sim}74.9$~million total). Each norm was fit independently using lme4 (v1.1-38; R v4.4.2)~\citep{bates2015lme4}.

To test whether observed idiosyncrasy variance $\sigma^2_{\iota}$ exceeds chance, we conducted a parametric bootstrap ($N = 100$). Each iteration generated data without an interaction term and refit the LMM; $p$-values reflect the proportion of null variances meeting or exceeding the observed value.

Cross-norm coherence was assessed via Ridge regression (5-fold cross-validation), predicting each model's BLUPs on one held-out norm from its BLUPs on the remaining 13 (within-model $R^2$), and comparing these to cross-model predictions (mean pairwise $R^2$) to yield a Specificity Ratio. Human alignment was assessed by correlating mean stochastic ratings with published human norms (Pearson $r$, Fisher-$z$ aggregated; see \emph{SI Appendix}).

The analysis code is available at \href{https://github.com/valentinkm/MachineIndividuality}{https://github.com/valentinkm/MachineIndividuality}. Claude (Anthropic) was used to assist with revising text for clarity. All scientific content, analyses, and interpretations are the sole responsibility of the authors.

% \section*{Acknowledgments}
% Add acknowledgments here if needed.

\bibliographystyle{unsrtnat}
\bibliography{references}

% SI Appendix — body-only fragment, to be \input from main.tex
% Auto-generated by build_arxiv.py from si.qmd — do not edit manually

\clearpage
\appendix
\sloppy

\setcounter{section}{0}
\renewcommand{\thesection}{S\arabic{section}}
\renewcommand{\thesubsection}{S\arabic{section}.\arabic{subsection}}
\renewcommand{\theequation}{S\arabic{equation}}
\renewcommand{\theHequation}{S\arabic{equation}}
\setcounter{equation}{0}

\begin{center}
{\LARGE\bfseries SI Appendix}\\[6pt]
{\large Machine individuality: Separating genuine idiosyncrasy from response bias in large language models}\\[6pt]
{Valentin Kriegmair, Dirk U. Wulff}
\end{center}

\vspace{1em}

\section{Models}\label{sec-models}

We selected 10 open-weight large language models spanning a broad range
of architectural implementations and parameter scales. All models are
instruction-tuned variants; they were accessed via offline vLLM
inference on NVIDIA H200 GPUs (141 GB HBM each).

\textbf{Qwen3-32B} (Alibaba; 32B parameters; \path{Qwen/Qwen3-32B}).
Dense transformer. Qwen3 dense models ship with seamless switching
between thinking and non-thinking modes; thinking was disabled at
inference via the \texttt{enable\_thinking=False} chat-template flag to
ensure direct numeric responses rather than extended chain-of-thought
reasoning that would interfere with response parsing.

\textbf{Qwen3-235B-A22B} (Alibaba; 235B total / 22B active parameters;
\path{Qwen/Qwen3-235B-A22B-Instruct-2507}). Mixture-of-experts (MoE)
architecture. The largest model in the sample by total parameter count.
Thinking disabled as above.

\textbf{Mistral-Small-24B-Instruct-2501} (Mistral AI; 24B parameters;
\path{mistralai/Mistral-Small-24B-Instruct-2501}). Dense transformer.

\textbf{gemma-3-27b-it} (Google; 27B parameters;
\path{google/gemma-3-27b-it}). Dense decoder-only transformer with
alternating local (sliding-window) and global self-attention. Deployed
with an exponential-backoff retry wrapper to mitigate transient vLLM
engine errors (max 5 retries, initial backoff 2 s).

\textbf{gpt-oss-20b} (OpenAI; 21B total / 3.6B active parameters;
\path{openai/gpt-oss-20b}). Mixture-of-experts transformer, shipped
with native MXFP4 quantization on the MoE weights (BF16 elsewhere). The
smallest model in the sample by active parameter count. Configured with
low reasoning effort to ensure direct numeric responses rather than
extended chain-of-thought reasoning.

\textbf{gpt-oss-120b} (OpenAI; 117B total / 5.1B active parameters;
\path{openai/gpt-oss-120b}). Larger variant of the OpenAI open-weight
family, sharing the same MoE architecture and native MXFP4 quantization.
Same reasoning configuration as gpt-oss-20b.

\textbf{OLMo-3.1-32B-Instruct} (Allen AI; 32B parameters;
\path{allenai/Olmo-3.1-32B-Instruct}). Dense transformer. Fully open
(weights, training data, and training code).

\textbf{Falcon-H1-34B-Instruct} (TII; 34B parameters;
\path{tiiuae/Falcon-H1-34B-Instruct}). Hybrid-head architecture
combining Transformer attention with Mamba-2 state-space heads in
parallel within each block.

\textbf{granite-4.0-h-small} (IBM; 32B total / 9B active parameters;
\texttt{ibm-granite/\mbox{granite-4.0-h-small}}). Hybrid Mamba-2/transformer
architecture with a fine-grained mixture-of-experts feed-forward block
(shared experts always active).

\textbf{phi-4} (Microsoft; 14B parameters; \path{microsoft/phi-4}).
Dense decoder-only transformer trained with a synthetic-data curriculum;
the smallest model in the sample by total parameter count.

Models were selected to sample broadly along three axes: (a)
organization (7 distinct providers), (b) architecture (dense
transformers, mixture-of-experts, and hybrid state-space models), and
(c) parameter scale (9B--235B total parameters, 9B--120B active
parameters). The sample deliberately favors recent releases over
established families (e.g., Llama) to capture the current frontier of
open-weight model diversity.

\section{Psycholinguistic Norms}\label{sec-norms}

We administered 14 psycholinguistic norms drawn from the
\emph{psychNorms} metabase \citep{hussain2024probingcontentssemanticrepresentations} to each model. The total
vocabulary comprised 107,083 cue words. Norms were selected to span a
broad range of evaluative dimensions --- from core affective and
perceptual judgments to higher-order social and moral appraisals. Below,
we describe each norm's target construct, scale, and human norming
sample.

\textbf{Arousal} \citep{Warriner2013}. Ratings were collected on a
9-point scale. The Warriner et al.~data-collection instructions
presented arousal from ``excited'' to ``calm'' (i.e., 1 = excited, 9 =
calm), which is the reverse of the original convention; the published
norms (and the values curated in psychNorms) were recoded back to the
ANEW direction (1 = calm, 9 = excited) in the erratum. Our LLM prompt
used the data-collection wording (1 = excited, 9 = calm), and model
ratings are reflected (\texttt{10\ -\ x}) to align with the recoded
human scale. Human norms: 13,915 English lemmas, collected from 1,827
U.S. Amazon Mechanical Turk workers. After psychNorms filtering, 13,784
words remained.

\textbf{Concreteness} \citep{Brysbaert2014}. Concreteness evaluates the
degree to which a word denotes a perceptible entity --- one that can be
experienced directly through the senses or motor actions --- versus an
abstract concept understood primarily through language. The construct is
central to Paivio's dual-coding theory. Ratings were collected on a 1--5
scale (1 = very abstract, 5 = very concrete). Human norms: 37,056 words,
each rated by at least 25 participants via Amazon Mechanical Turk.

\textbf{Valence} \citep{Mohammad2018}. The valence norms from the NRC
Valence, Arousal, and Dominance Lexicon \citep{Mohammad2018} provide
real-valued scores in [0, 1] for 19,971 English terms. Ratings were
derived via best--worst scaling (BWS): annotators saw tuples of four
words and indicated the most and least positively valenced in each
tuple. Per-word scores were then computed as the proportion of times a
word was chosen as ``most positive'' minus the proportion it was chosen
as ``least positive,'' aggregated across all tuples. This yields a
relative ranking on a continuous scale rather than Likert-style absolute
ratings. Collection involved 1,020 crowdworkers and six annotations per
tuple. Because BWS scores are not directly elicitable from a single-item
prompt, we prompted models to rate positivity and negativity separately
on two unipolar 0--3 scales and combined them into a signed bipolar
score in [-3, +3]. 19,382 words are retained in psychNorms.

\textbf{Sensory norms: Visual, Auditory, Gustatory, Olfactory, and
Haptic} \citep{Lynott2020}. The Lancaster Sensorimotor Norms quantify
the perceptual strength of word meanings across sensory modalities,
capturing the degree to which a concept is experienced through a
specific sensory channel. We used five of the six perceptual modalities
(excluding interoceptive) and none of the five action effector
dimensions. Ratings were collected on a 0--5 scale (0 = not experienced
at all, 5 = experienced greatly). Human norms: 36,808 words per
modality, rated by 2,625 participants via Amazon Mechanical Turk, with
approximately 19 ratings per word.

\textbf{Age of Acquisition --- Kuperman} \citep{Kuperman2012}.
Subjective AoA estimates. Participants typed the estimated age in years
at which they first understood each word. Human norms: 30,121 English
content words, collected from 1,960 Amazon Mechanical Turk workers. The
psychNorms entry reflects subsequent erratum updates and contains 30,873
lemmas.

\textbf{Age of Acquisition --- Brysbaert} \citep{Brysbaert2017}.
Test-based AoA derived from Dale and O'Rourke's \emph{Living Word
Vocabulary} \citep{Dale1981TheLW}. We elicited grade-level responses
from \{4, 6, 8, 10, 12, 13, 16\}. In the original study, U.S. children
were given three-alternative multiple-choice meaning-recognition items
at grades 4, 6, 8, 10, 12, and two college levels (13 and 16), with a
meaning assigned to a grade when 67--80\% of $\sim$200 children
passed --- corresponding, after correction for guessing on a
three-alternative item, to roughly 50--70\% actually knowing the meaning
\citep{Brysbaert2017}. 19,953 word-forms are retained in psychNorms.

\textbf{Morality} \citep{Troche2017}. Each noun was rated on 7-point
Likert scales indicating agreement with dimension-specific statements;
the morality item asked participants to rate agreement with the
statement that morality is central to the concept. Human norms: 750
English nouns drawn from the MRC Psycholinguistic Database, rated by N =
328 native-English Amazon Mechanical Turk workers (mean age 34.63). 751
items are curated in psychNorms.

\textbf{Gender Association} \citep{Scott2019}. Gender association
measures the degree to which a word's meaning is associated with male or
female behavior, extending beyond gender-specific nouns to encompass
objects, actions, and abstract concepts carrying gendered connotations.
It is one of nine dimensions in the Glasgow Norms. Ratings were
collected on a 1--7 scale (1 = very feminine, 4 = neuter, 7 = very
masculine). Human norms: 4,668 words, rated by 829 participants from the
University of Glasgow community, with approximately 33 ratings per word.

\textbf{Humor} \citep{Engelthaler2018}. Humor captures the perceived
funniness of individual words and has been shown to be only weakly
correlated with other psycholinguistic dimensions (arousal, valence,
dominance, concreteness), suggesting it captures a largely independent
aspect of word meaning. Ratings were collected on a 1--5 scale (1 =
humorless, 5 = humorous). Human norms: 4,996 words, rated by 821
participants via Amazon Mechanical Turk.

\textbf{Socialness} \citep{Diveica2023}. Socialness rates the degree to
which a word's meaning has social relevance, using an inclusive
definition encompassing social characteristics, behaviors, roles,
spaces, institutions, values, and other socially relevant constructs.
The construct is motivated by multiple-representation theories of
semantic memory, which propose social experience as a fundamental source
of conceptual meaning. Ratings were collected on a 1--7 scale. Human
norms: 8,387 words, rated by 605 participants via Prolific, with at
least 25 ratings per word.

\section{Prompt Templates}\label{sec-prompts}

Models received zero-shot prompts adapted from the original studies'
human rating instructions. Each prompt describes the rating task and
scale, presents the target word, and instructs the model to respond with
a single number. The placeholder \texttt{\{word\}} is replaced by the
target cue word at elicitation time.

\subsection[Arousal]{Arousal
\citep{Warriner2013}}\label{arousal-warriner2013}

\begin{verbatim}
You are invited to take part in a study that is investigating emotion,
and concerns how people respond to different types of words. You will
use a scale to rate how you felt while reading each word.

The scale ranges from 1 (excited) to 9 (calm). At one extreme of this
scale, you are stimulated, excited, frenzied, jittery, wide-awake, or
aroused. When you feel completely aroused you should indicate this by
choosing rating 1. The other end of the scale is relaxed, calm,
sluggish, dull, sleepy, or unaroused. You can indicate feeling
completely calm by selecting 9. The numbers also allow you to describe
intermediate feelings of calmness/arousal. If you feel completely
neutral -- not excited nor at all calm -- select the middle of the
scale (rating 5).

Word: "{word}"

Your response MUST start with a single number from 1 to 9 and contain
nothing else.
Rating:
\end{verbatim}

\subsection[Concreteness]{Concreteness
\citep{Brysbaert2014}}\label{concreteness-brysbaert2014}

\begin{verbatim}
You are participating in a psychology experiment. Your task is to rate
a word based on its concreteness using a 5-point scale.

1: very abstract
2: abstract
3: in-between
4: concrete
5: very concrete

Word: "{word}"

Your response MUST start with a single number from 1 to 5 and contain
nothing else.
Rating:
\end{verbatim}

\subsection[Valence --- Positive]{Valence --- Positive
\citep{Mohammad2018}}\label{valence-positive-mohammad2018}

\begin{verbatim}
You are participating in a psychology experiment. Your task is to rate
how positive (good, praising) the word is. Word: "{word}"
Scale (0-3):
0: "{word}" is not positive
1: "{word}" is weakly positive
2: "{word}" is moderately positive
3: "{word}" is strongly positive

Your response MUST start with a single number from 0 to 3 and contain
nothing else.
Rating:
\end{verbatim}

\subsection[Valence --- Negative]{Valence --- Negative
\citep{Mohammad2018}}\label{valence-negative-mohammad2018}

\begin{verbatim}
You are participating in a psychology experiment. Your task is to rate
how negative (bad, criticizing) the word is.

Word: "{word}"
Scale (0-3):
0: "{word}" is not negative
1: "{word}" is weakly negative
2: "{word}" is moderately negative
3: "{word}" is strongly negative

Your response MUST start with a single number from 0 to 3 and contain
nothing else.
Rating:
\end{verbatim}

\subsection[Sensory Norms]{Sensory Norms
\citep{Lynott2020}}\label{sensory-norms-lynott2020}

For each of the five perceptual modalities, the prompt follows the same
structure. Below is the Visual template as an example; the other four
substitute the relevant modality (Auditory/hearing, Gustatory/tasting,
Olfactory/smelling, Haptic/touch).

\begin{verbatim}
You are participating in a psychology experiment. Your task is to rate
to what extent you experience a concept through sight.

Scale (0-5): 0 = Not at all, 5 = To a great extent

Word: "{word}"

Your response MUST start with a single number from 0 to 5 and contain
nothing else.
Rating:
\end{verbatim}

\subsection[AoA --- Kuperman]{AoA --- Kuperman
\citep{Kuperman2012}}\label{aoa-kuperman-kuperman2012}

\begin{verbatim}
You are participating in a linguistics experiment. Your task is to
estimate the age (in years) at which an average person first learned
a word (understood it when heard).

Word: "{word}"

Your response MUST be a single integer number representing the age and
contain nothing else.
Age:
\end{verbatim}

\subsection[AoA --- Brysbaert]{AoA --- Brysbaert
\citep{Brysbaert2017}}\label{aoa-brysbaert-brysbaert2017}

\begin{verbatim}
You are participating in a vocabulary experiment. Estimate the grade
level at which an average student would know this word in a
three-choice vocabulary test, defined as the first grade where at
least 50% answer correctly (corrected for guessing).

Choose one of these grade levels only: 4, 6, 8, 10, 12, 13, or 16.

Word: "{word}"

Your response MUST be a single number from 4, 6, 8, 10, 12, 13, or 16
and contain nothing else.
Grade:
\end{verbatim}

\subsection[Morality]{Morality
\citep{Troche2017}}\label{morality-troche2017}

\begin{verbatim}
You are participating in a psychology experiment. Your task is to
indicate how much you agree with the statement: "I relate this word to
morality, rules or any other thing that governs my behavior."

Scale (1-7):
1: Strongly disagree
2: Disagree
3: Somewhat disagree
4: Neutral
5: Somewhat agree
6: Agree
7: Strongly agree

Word: "{word}"

Your response MUST start with a single number from 1 to 7 and contain
nothing else.
Rating:
\end{verbatim}

\subsection[Gender Association]{Gender Association
\citep{Scott2019}}\label{gender-association-scott2019}

\begin{verbatim}
You are participating in a psychology experiment. Your task is to rate
the gender association of the word.

A word's gender is how strongly its meaning is associated with male or
female behaviour. A word can be considered MASCULINE if it is linked to
male behaviour. Alternatively, a word can be considered FEMININE if it
is linked to female behaviour. Please indicate the gender associated
with each word on a scale of VERY FEMININE to VERY MASCULINE, with the
midpoint being neuter (neither feminine nor masculine).

Scale (1-7): 1 = Very feminine, 4 = Neuter, 7 = Very masculine

Word: "{word}"

Your response MUST start with a single number from 1 to 7 and contain
nothing else.
Rating:
\end{verbatim}

\subsection[Humor]{Humor
\citep{Engelthaler2018}}\label{humor-engelthaler2018}

\begin{verbatim}
You are participating in a psychology experiment. You will rate how you
feel while reading each word on a humor scale.

The scale ranges from 1 (humorless = not funny at all) to 5
(humorous = most funny). If you find the word dull or unfunny, give it
a rating of 1. If you feel the word is amusing or likely to be
associated with humorous thought or language (e.g., absurd, amusing,
hilarious, playful, silly, whimsical, or laughable), give it a rating
of 5. Use intermediate numbers for words that fall between these
extremes. If the word is neutral (neither humorous nor humorless),
select the middle of the scale (rating 3).

Word: "{word}"

Your response MUST start with a single number from 1 to 5 and contain
nothing else.
Rating:
\end{verbatim}

\subsection[Socialness]{Socialness
\citep{Diveica2023}}\label{socialness-diveica2023}

\begin{verbatim}
You are participating in a psychology experiment. Your task is to rate
the degree to which the word's meaning has social relevance by
describing or referring to: a social characteristic of a person or
group of people, a social behaviour or interaction, a social role, a
social space, a social institution or system, a social value or
ideology, or any other socially relevant concept.

Scale (1-7)

Word: "{word}"

Your response MUST start with a single number from 1 to 7 and contain
nothing else.
Rating:
\end{verbatim}

\section{Data Collection \& Exclusion}\label{sec-data}

\subsection{Prompting and generation
procedure}\label{prompting-and-generation-procedure}

Each rating was elicited as a single-turn, zero-shot conversation: the
prompt template (Section~\ref{sec-prompts}) was formatted with the target word
and sent as a user message to the model's chat endpoint. Each model was
deployed using the vLLM offline inference engine on NVIDIA H200 GPUs
(141 GB HBM each) on the MPCDF DAIS cluster.

\textbf{Generation parameters.} All models shared the following default
parameters: a maximum of 256 output tokens and a newline stop sequence
(remaining vLLM defaults: top-\(p\) = 1.0, repetition penalty = 1.0,
context window capped at 4,096 tokens, GPU memory utilization = 0.90).
The two GPT-OSS models used an extended output limit of 4,096 tokens to
accommodate their internal reasoning chain before the final numeric
answer. For stochastic ratings, temperature was set to \(T = 1.0\) with
5 independent repetitions per cue word; for deterministic ratings,
\(T = 0\) with a single repetition. Two Qwen models additionally
disabled internal thinking via a chat template override to suppress
chain-of-thought tokens.

\textbf{Response parsing.} Raw model outputs were cleaned by a
model-specific adapter: any internal thinking blocks were stripped (Qwen
models), and the leading numeric value was extracted via regex.
Responses that could not be parsed to a valid number were marked as
unparseable.

\textbf{Multi-stage retry logic.} Failed responses underwent up to 3
retry attempts across 4 sequential stages:

\begin{itemize}
\item
  \emph{Stage 1 (Scale)}: If the extracted number fell outside the
  norm's valid range, the original prompt was augmented with an explicit
  scale reminder (``Your previous answer was invalid. Please output a
  single number between \emph{min} and \emph{max}.'').
\item
  \emph{Stage 2 (Parse)}: Unparseable responses were re-submitted with
  the original prompt unchanged.
\item
  \emph{Stage 3 (Temperature)}: Remaining parse and scale failures were
  re-submitted with a slightly elevated temperature (+0.1) to perturb
  the output distribution.
\item
  \emph{Stage 4 (Refusal)}: Responses matching refusal patterns (e.g.,
  ``as an AI,'' ``I cannot'') were re-submitted with one of two
  augmented prompts, depending on whether the refusal was classified as
  a safety refusal (scientific-context framing) or a generic refusal
  (role-play constraint).
\end{itemize}

\subsection{Postprocessing and
exclusion}\label{postprocessing-and-exclusion}

Raw model outputs were postprocessed by a unified pipeline:

\begin{enumerate}
\def\labelenumi{\arabic{enumi}.}
\item
  \textbf{Rating extraction.} Free-form text responses were parsed to
  extract the leading numeric value. Responses that could not be parsed
  to a valid number were marked as invalid.
\item
  \textbf{Scale validation.} Extracted ratings were validated against
  each norm's defined scale. Out-of-range values were flagged as
  outliers.
\item
  \textbf{Bipolar valence combination.} For the valence norm, the
  positive rating (0--3) minus the negative rating (0--3) was computed
  per (model, word, repetition) pair, yielding a single bipolar score in
  [-3, +3].
\item
  \textbf{Repetition cap.} Stochastic repetitions were capped at 5 per
  (model, norm, word) tuple.
\item
  \textbf{Effective validity.} An observation was marked as effectively
  valid if it was parseable, within scale, and not flagged by any prior
  exclusion step.
\end{enumerate}

After exclusion, the overall invalid/outlier rate was low ($<$3\%
per norm, mean $<$0.5\%), and approximately 5.4 million valid
observations per norm entered the LMM analysis ($\sim$74.9
million total across 14 norms).

\section{Variance Decomposition --- Model Specification}\label{sec-lmm}

Each of the 14 norms was fit independently using a crossed
random-effects linear mixed model via \texttt{lme4} \citep{bates2015lme4}
(v1.1-38; R v4.4.2):

\begin{equation}
y_{ijk} = \mu + \tau_i + \beta_j + \iota_{ij} + \epsilon_{ijk}
\end{equation}

where \(y_{ijk}\) is the rating for word \(i\), model \(j\), and
repetition \(k\); \(\mu\) is the grand mean;
\(\tau_i \sim \mathcal{N}(0, \sigma^2_\tau)\) captures shared consensus
across models (Trait); \(\beta_j \sim \mathcal{N}(0, \sigma^2_\beta)\)
captures systematic directional model bias (Bias);
\(\iota_{ij} \sim \mathcal{N}(0, \sigma^2_\iota)\) captures the
word$\times$model interaction (Idiosyncrasy); and
\(\epsilon_{ijk} \sim \mathcal{N}(0, \sigma^2_\epsilon)\) is residual
noise.

Models were fit using the BOBYQA optimizer (lme4 defaults; derivative
calculations disabled). Variance proportions were computed as the ratio
of each component's variance to the total variance
(\(\sigma^2_\tau + \sigma^2_\beta + \sigma^2_\iota + \sigma^2_\epsilon\)).
For the aggregated dimensions reported in Fig. 1A, norms were grouped as
follows: the five sensory norms into ``Sensory Norms'' and the two AoA
norms into ``Age of Acquisition''; all other norms formed singleton
dimensions. The aggregated proportions are the arithmetic mean of the
individual norm-level variance proportions within each group (not a
pooled refit across the combined observations).

\section{Parametric Null Simulation}\label{sec-simulation}

To test whether the observed idiosyncrasy variance \(\sigma^2_\iota\)
exceeds what could arise from structural artefacts of the crossed design
(i.e., spurious interaction variance), we conducted a parametric
bootstrap (\(N = 100\) iterations per norm). Each iteration proceeded as
follows:

\begin{enumerate}
\def\labelenumi{\arabic{enumi}.}
\item
  Drew fresh random effects
  \(\tilde{\tau}_i \sim \mathcal{N}(0, \hat{\sigma}^2_\tau)\) and
  \(\tilde{\beta}_j \sim \mathcal{N}(0, \hat{\sigma}^2_\beta)\) from the
  estimated distributions fitting the empirical model.
\item
  Generated synthetic data \emph{without} an interaction term:
  \(\tilde{y}_{ijk} = \hat{\mu} + \tilde{\tau}_i + \tilde{\beta}_j + \epsilon_{ijk}\),
  where \(\epsilon_{ijk} \sim \mathcal{N}(0, \hat{\sigma}^2_\epsilon)\).
\item
  Re-fit the full model (including the interaction term \(\iota_{ij}\))
  to the simulated data.
\item
  Recorded the spurious interaction variance
  \(\sigma^2_{\iota,\text{null}}\).
\end{enumerate}

The \(p\)-value for each norm is the proportion of null iterations where
\(\sigma^2_{\iota,\text{null}} \geq \sigma^2_{\iota,\text{observed}}\).

Each simulation iteration was run on the full dataset
($\sim$5.4M observations per norm across 10 models), preserving
the complete crossed design without subsampling. Simulations were
parallelized across norms (14 concurrent workers, 2 cores each) on a
128-core, 500 GB RAM Linux server.

\section{Cross-Norm Specificity}\label{sec-specificity}

To assess whether model idiosyncrasies form coherent, model-specific
fingerprints rather than isolated norm-level fluctuations, we used Ridge
regression with 5-fold cross-validation. For each model \(j\) and each
held-out norm \(k\):

\begin{enumerate}
\def\labelenumi{\arabic{enumi}.}
\item
  The best linear unbiased predictions (BLUPs) \(\hat{\iota}_{ij}\) on
  the held-out norm served as the target vector (one value per word in
  the shared vocabulary).
\item
  The same model's BLUPs on the remaining 13 norms served as the
  predictor matrix.
\item
  A \emph{within-model} \(R^2\) was computed via 5-fold cross-validated
  Ridge regression (regularization selected by internal CV).
\item
  \emph{Cross-model} \(R^2\) values were computed by predicting model
  \(j\)'s BLUPs on the held-out norm from each other model \(j'\)'s
  BLUPs on the same 13 predictor norms.
\item
  The Specificity Ratio = within-model \(R^2\) / mean cross-model
  \(R^2\).
\end{enumerate}

A Specificity Ratio $>$ 1 indicates that a model's deviations
on one dimension are better predicted by its own deviations on other
dimensions than by those of other models. The same analysis was applied
to raw ratings (without removing trait and bias components) to test
whether the specificity signal is carried specifically by the
idiosyncrasy component.

The shared vocabulary for the Ridge regression was determined per model
as the intersection of word sets across all 14 norms' BLUP files (i.e.,
words for which valid BLUPs were available on every norm for that
model).

To prevent BLAS multi-threading from interfering with the outer
parallelization loop, all threads were pinned to a single core per
worker.

\section{Human Correlation}\label{sec-human}

Human benchmark norms were drawn from the \emph{psychNorms} metabase
\citep{hussain2024probingcontentssemanticrepresentations}, which aggregates published psycholinguistic norms
into a unified, word-indexed dataset. For each of the 14 norms, the
metabase provides the aggregated human rating per word from the original
norming study (see Section~\ref{sec-norms} for sample sizes and per-norm
details). Correlations were computed on the intersection of the model's
vocabulary and the words available in the human norming study for each
norm; overlap sizes range from 751 words (morality) to 37,056 words
(concreteness), reflecting the varying coverage of the original human
studies.

For each model and each norm, we computed the Pearson correlation \(r\)
between the model's mean stochastic rating (averaged over 5 repetitions
per word) and the corresponding published human rating. The correlation
was computed on raw (unstandardized) values; because Pearson \(r\) is
invariant to affine transformations, differences in scale ranges between
model and human data (e.g., model valence on \([-3, +3]\) vs.~human
valence on \([0, 1]\)) do not affect the correlation coefficient.

\textbf{Arousal scale convention.} The arousal prompt reproduces the
original Warriner et al.~collection instructions, in which 1 = excited
and 9 = calm. The published human norms were recoded post-hoc by
Warriner et al.~to a more intuitive low-to-high direction (1 = calm, 9 =
excited) \citep{Warriner2013}. To align directions before correlating,
model arousal ratings are reversed (\(10 - x\)) at analysis time,
matching the convention of the published norms.

To aggregate across norms, we applied the Fisher \(z\)-transformation:
each per-norm \(r\) was transformed to \(z = \text{arctanh}(r)\), the
\(z\)-values were averaged across the 14 norms, and the aggregate was
back-transformed to \(\bar{r} = \tanh(\bar{z})\). This yields a single
model-level human alignment score.

For each (model, norm) pair, we computed Pearson correlations against
published human norms separately for (a) the stochastic mean (average of
5 repetitions at \(T = 1.0\)) and (b) the deterministic rating
(\(T = 0\), single response). The stochastic advantage is defined as
\(\Delta r = r_{\text{stochastic mean}} - r_{\text{deterministic}}\),
quantifying the human alignment gained by aggregating across stochastic
repetitions relative to a single deterministic response.

\section{Software and Reproducibility}\label{sec-software}

\subsection{Data generation}\label{data-generation}

Models were deployed via offline vLLM inference on the DAIS cluster at
the Max Planck Computing and Data Facility (MPCDF). Each node is
equipped with 8 NVIDIA H200 GPUs (141 GB HBM each), 96 CPU cores, and 2
TB RAM.

\subsection{Analysis pipeline}\label{analysis-pipeline}

The analysis pipeline was executed on a 128-core, 500 GB RAM Linux
server. A master orchestration script coordinates 9 sequential steps
from postprocessing through publication figure generation.

\textbf{Software versions:}

\begin{itemize}
\item
  Python 3.11, pandas \(\geq\) 2.1, numpy \(\geq\) 1.26, scipy \(\geq\)
  1.12, scikit-learn \(\geq\) 1.4, pyarrow \(\geq\) 15.0
\item
  R 4.4.2, lme4 \citep{bates2015lme4} v1.1-38, data.table \(\geq\) 1.15,
  arrow \(\geq\) 15.0
\end{itemize}

A full environment specification for conda/micromamba is provided in the
repository. All analysis code and supporting datasets are available at
\url{https://github.com/valentinkm/MachineIndividuality}. The raw
behavioral data of all models can be found on OSF:
\url{https://doi.org/10.17605/OSF.IO/T9S3M}.

\end{document}